\documentclass[10pt,conference]{IEEEtran}
\usepackage{amsmath,amssymb,amsfonts}
\usepackage{graphicx}
\usepackage{booktabs}
\usepackage{hyperref}
\usepackage{multirow}
\usepackage{xcolor}
\usepackage{threeparttable}

\newcommand\blfootnote[1]{%
  \begingroup
  \renewcommand\thefootnote{}\footnote{\rule{0.5\textwidth}{0.4pt}\\#1}%
  \addtocounter{footnote}{-1}%
  \endgroup
}

\usepackage{titlesec}
\titleformat*{\section}{\normalfont\Large\bfseries}
\titleformat*{\subsection}{\normalfont\large\bfseries}
\titleformat*{\subsubsection}{\normalfont\large\bfseries}

\begin{document}

\title{Spatial ModernBERT: Spatial-Aware Transformer for Table and Key-Value Extraction in Financial Documents at Scale}

\author{
   \IEEEauthorblockN{\textbf{Amrendra Singh}\\{amrendra.singh@javis.ai}}
   \and
   \IEEEauthorblockN{\textbf{Maulik Shah}\\{maulik.shah@javis.ai}}
   \vspace{1.0em}
   \IEEEauthorblockA{\textbf{Javis AI Team}}
   \small{
      \textit{* work done while part of the Javis AI Team}
   }
   \and
   \IEEEauthorblockN{\textbf{Dharshan Sampath}\textsuperscript{*}}
}

\maketitle

\begin{abstract}
Extracting tables and key-value pairs from financial documents is essential for business workflows such as auditing, data analytics, and automated invoice processing. In this work, we introduce Spatial ModernBERT—a transformer-based model augmented with spatial embeddings—to accurately detect and extract tabular data and key-value fields from complex financial documents. We cast the extraction task as token classification across three heads: (1) Label Head, classifying each token as a label (e.g., PO Number, PO Date, Item Description, Quantity, Base Cost, MRP, etc.); (2) Column Head, predicting column indices; (3) Row Head, distinguishing the start of item rows and header rows. The model is pretrained on the PubTables-1M dataset, then fine-tuned on a financial document dataset, achieving robust performance through cross-entropy loss on each classification head. We propose a post-processing method to merge tokens using B-I-IB tagging, reconstruct the tabular layout, and extract key-value pairs. Empirical evaluation shows that Spatial ModernBERT effectively leverages both textual and spatial cues, facilitating highly accurate table and key-value extraction in real-world financial documents.
\end{abstract}


\blfootnote{Code available at: \url{https://github.com/javis-admin/Spatial-ModernBERT}}

\section{Introduction}
Financial documents such as invoices, purchase orders, and receipts often contain critical information in the form of tables (item descriptions, pricing, taxes, etc.) and key-value pairs (PO Number, PO Date, Vendor Name, etc.). Automating the extraction of this data saves immense manual effort and reduces errors in auditing and business intelligence tasks. However, table structures can be highly variable, with multi-column layouts, embedded text, and differing row structures. Traditional rule-based methods often fail on such diverse layouts, and recent advances in deep learning provide an opportunity to handle these variations more systematically.

Prior approaches to table extraction from documents can be broadly categorised into three major categories:

\begin{enumerate}
\item \textbf{Object Detection Approaches}: 
   These computer vision-based methods identify table elements as visual objects. They are computationally efficient and work well for tables with clear boundaries, but struggle with complex layouts and require extensive retraining for new document formats. Models in this category include Faster R-CNN \cite{faster-rcnn}, TableNet \cite{tablenet}, DeepDeSRT \cite{deepdesrt}, and CascadeTabNet \cite{cascadetabnet}.

\item \textbf{Decoder-Only Models}: 
   These LLM-based approaches generate structured table representations directly from document text and visual features. While they excel at understanding semantic relationships and handling diverse layouts, they are computationally expensive, slow during inference, and prone to hallucination. Examples include Donut \cite{donut}, GPT, Gemini, Claude, Qwen \cite{qwen}, and LLaVA \cite{llava}-based models.

\item \textbf{Two-Stage Detection and Relationship Extraction}: 
   Methods like Res2Tim \cite{res2tim}, TableFormer \cite{tableformer}, FLAG-Net \cite{flag-net}, and GraphTSR \cite{graphtsr} first identify individual cells and then determine relationships between them. These approaches handle complex tables with irregular layouts but introduce computational overhead and depend on additional algorithmic layers.
\end{enumerate}

In this paper, we introduce Spatial ModernBERT, a transformer-based system that treats table extraction as a multi-headed token classification task. Our contributions are summarised as follows:

\begin{enumerate}
\item We propose a novel multi-headed token classification approach for table and key-value extraction that simultaneously learns semantic labels, tabular structure (rows and columns), and spatial relationships. Our approach achieves an F1 score of 95.49 on CORD and 96.91\% TEDS on PubTabNet, demonstrating superior performance over existing methods.

\item We leverage the spatial embedding scheme from LayoutLMv3 \cite{layoutlmv3} that effectively captures document layout information by embedding both 2D-positional coordinates and dimensional properties of text tokens. This 768-dimensional spatial embedding combines $x_{\min}$, $y_{\min}$, $x_{\max}$, $y_{\max}$, $width$, and $height$ information into a rich representation that improves spatial learning.

\item We develop a specialised B-I-IB tagging scheme that handles multi-line text fields commonly found in complex financial documents. This tagging innovation enables the model to maintain semantic continuity across line breaks, improving multi-line field extraction compared to traditional BIO tagging.

\item We present a comprehensive ablation study analysing the impact of various architectural components, including different spatial embedding configurations and initialisations, loss weighting strategies, additional losses, and training objectives.

\item We demonstrate computational efficiency with near-linear scaling up to batch size 8 and consistent throughput across longer sequence lengths, processing up to 60,000 tokens per second, making our approach suitable for high-volume processing in production environments.
\end{enumerate}

\section{Related Work}

\subsection{Table Extraction}
Early approaches to table extraction often relied on rule-based heuristics or classical computer vision to locate horizontal and vertical rulings. These methods typically use line detection algorithms (e.g., Hough transform \cite{hough}) to identify table boundaries and cell separators. With the advent of large-scale annotated datasets, machine learning-based approaches have significantly outperformed rule-based methods.

Convolutional Neural Networks (CNNs) were among the first deep learning models applied to table detection. TableNet \cite{tablenet} utilised a dual-branch CNN architecture to simultaneously segment table regions and column regions. DeepDeSRT \cite{deepdesrt} extended this by incorporating a region proposal network to identify possible table regions first, followed by structure recognition. CascadeTabNet \cite{cascadetabnet} introduced a cascade mask R-CNN approach with a novel image processing pipeline for handling both bordered and borderless tables. 

More recently, transformer-based models have achieved state-of-the-art performance. DETR \cite{detr} (DEtection TRansformer) has been adapted for table detection by treating tables as objects in a visual scene. Other notable approaches include EDD \cite{edd} (Ensemble of Deep Detectors), which combines multiple detection models, and SEM \cite{sem} (Structure Extraction Method), which focuses on recognising the underlying logical structure of tables beyond mere detection.

Our use of a transformer backbone (ModernBERT) continues this trend, while specifically targeting financial document layouts and extending capabilities to simultaneously extract tabular data and key-value pairs.

\subsection{Key-Value Extraction}
Key-value extraction from documents has been explored widely in the context of information retrieval and document analysis. Initial approaches relied on pattern matching and rule-based systems that identified spatial relationships between text elements.

Named Entity Recognition (NER) techniques were subsequently adapted for document understanding, with models like SPADE \cite{spade} (Spatial Document Analysis) incorporating layout features to enhance text-based extraction. FUNSD \cite{funsd} (Form Understanding in Noisy Scanned Documents) introduced a benchmark dataset and approaches leveraging both textual and spatial features.

Transformer models have revolutionised this field. LayoutLM \cite{layoutlm} pioneered the integration of text, layout, and image information in a unified architecture for document understanding tasks, including key-value extraction. LayoutLMv2 \cite{layoutlmv2} enhanced this approach with visual embeddings and a spatial-aware self-attention mechanism. LayoutLMv3 \cite{layoutlmv3} further improved performance with pre-training objectives specifically designed for document intelligence tasks. Other notable models include BROS \cite{bros} (BERT Relying On Spatiality), which focuses on spatial relationships between text entities, and DocFormer \cite{docformer}, which uses multimodal attention mechanisms to fuse textual and visual information.

SelfDoc \cite{selfdoc} and TILT \cite{tilt} (Text-Image-Layout Transformer) have also demonstrated strong performance on key-value extraction tasks by effectively modelling the interactions between different modalities. XYLayoutLM \cite{xylayoutlm} introduced a novel approach to encoding 2D positions of text, further improving extraction accuracy. These approaches all share the common thread of leveraging both textual content and spatial information to identify entity-value relationships in document layouts.

Spatial ModernBERT builds upon these advances by incorporating textual embeddings with bounding box coordinates for better structured data extraction, while emphasising computational efficiency through our token classification approach.

\subsection{Spatially Aware Language Models}
Embedding spatial features directly into language models has emerged as a powerful approach for document understanding. The key insight is that document layout provides critical context that complements textual information.

LayoutLM \cite{layoutlm} was among the first to integrate spatial information by encoding bounding box coordinates of text segments into 2D-positional embeddings. This approach was refined in LayoutLMv2 \cite{layoutlmv2}, which added visual features and a spatial-aware self-attention mechanism to better model relationships between text regions. LayoutLMv3 \cite{layoutlmv3} further enhanced this with pre-training tasks specifically designed to learn spatial relationships.

LiLT \cite{lilt} (Language-independent Layout Transformer) addressed cross-lingual document understanding by separating text and layout streams, enabling transfer learning across languages. TILT \cite{tilt} similarly employed a dual-stream architecture but with a focus on text-layout alignment through contrastive learning objectives.

DiT \cite{dit} (Document Image Transformer) and BEiT \cite{beit} (Bidirectional Encoder representation from Image Transformers) adapted vision transformer architectures to incorporate document-specific spatial understanding. Donut \cite{donut} (Document understanding transformer) took a different approach by using an encoder-decoder architecture that jointly processes text and layout information for end-to-end document understanding.

More specialized models include StrucText \cite{structext}, which incorporates a structure-aware pre-training strategy, and UniDoc \cite{unidoc} (Unified Document Processing), which handles multiple document understanding tasks through a universal interface. Pix2Struct \cite{pix2struct} has demonstrated strong zero-shot capabilities by training on diverse document types with a focus on spatial reasoning.

Our approach continues this line of research by computing rich spatial features from bounding box coordinates and integrating them with text embeddings. Spatial ModernBERT's spatial embedding scheme combines not just 2D-positional information ($x_{\min}$, $y_{\min}$, $x_{\max}$, $y_{\max}$) but also dimensional properties ($width$, $height$) to capture richer layout cues, enabling more effective document understanding for table and key-value extraction tasks.

\begin{figure*}[t]
\centering
\includegraphics[width=\textwidth]{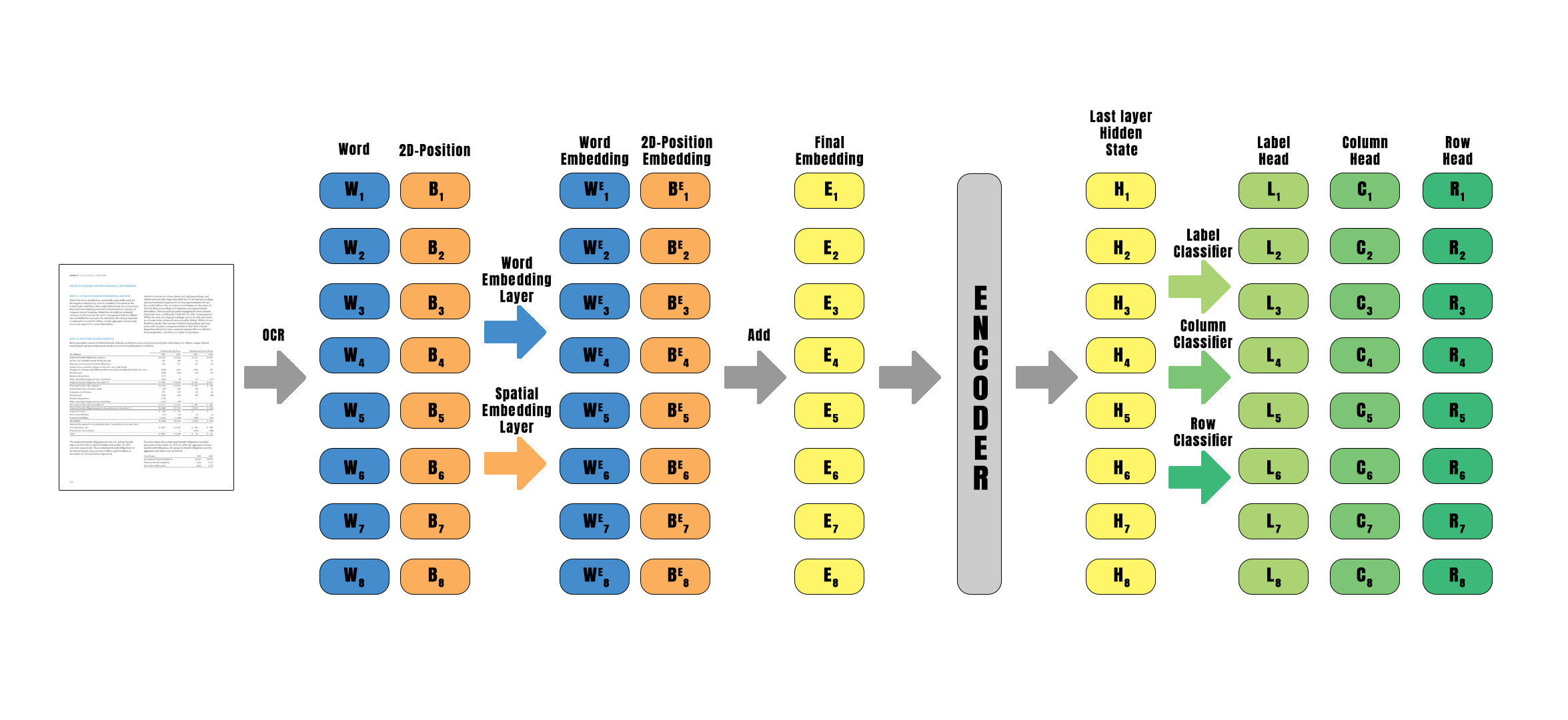}
\caption{Overall architecture of Spatial ModernBERT, showing the integration of text and spatial embeddings, transformer encoder, and multi-headed classification for table and key-value extraction.}\label{fig:architecture}
\end{figure*}

\section{Methodology}

\subsection{Overall Architecture}
Our system uses ModernBERT-base \cite{modernbert} as the base large language model, extended with custom spatial embeddings and three classification heads for table and key-value extraction, as shown in Fig.~\ref{fig:architecture}.

\subsubsection{Input Representation}
Each document is first processed to extract:
\begin{itemize}
\item Words: The text tokens from OCR or digital sources.
\item Bounding Box Coordinates ($x_{\min}$, $y_{\min}$, $x_{\max}$, $y_{\max}$) for each token in the range [0, 1000].
\end{itemize}

\subsubsection{Spatial Embeddings}
For each token, we transform bounding box data into a 768-dimensional embedding:
\begin{itemize}
\item We embed $x_{\min}$, $y_{\min}$, $x_{\max}$, $y_{\max}$ each into a 128-dimensional vector.
\item Additionally, we compute $width = x_{\max} - x_{\min}$ and $height = y_{\max} - y_{\min}$. Each is also mapped into a 128-dimensional vector.
\item We concatenate these six 128-dimensional vectors into a 768-dimensional spatial embedding.
\end{itemize}

The resulting spatial embedding is added to the standard ModernBERT \cite{modernbert} text embedding (768-dimensional) for each token, enabling the model to leverage both text and layout information.

\subsubsection{Spatial ModernBERT Backbone}
We use ModernBERT-base \cite{modernbert} as our text encoder, which is a transformer-based large language model. For each token, the combined text and spatial embeddings are fed into Spatial ModernBERT's encoder layers to produce a contextualized representation.

\subsubsection{Multi-Head Token Classification}
We append classification heads on top of the final encoder states:

\begin{enumerate}
\item \textbf{Label Head}
   \begin{itemize}
   \item Predicts the semantic label of each token (e.g., PO Number, PO Date, Item Description, Quantity, Base Cost, MRP, etc.).
   \item Includes B-I-IB tagging to handle multi-token labels that can span multiple lines.
   \end{itemize}

\item \textbf{Column Head}
   \begin{itemize}
   \item Assigns each token to a column index (0 through 9). Columns beyond 9 cycle back to 0.
   \item Labels include B-col\_0, I-col\_0, IB-col\_0, \ldots, B-col\_9, I-col\_9, IB-col\_9, plus an O label for tokens not in a tabular structure.
   \end{itemize}

\item \textbf{Row Head}
   \begin{itemize}
   \item Predicts whether a token belongs to an item row or header row.
   \item Labels include B-row, I-row, IB-row, B-header\_row, I-header\_row, IB-header\_row, and O.
   \end{itemize}
\end{enumerate}

\textbf{Tagging Scheme (B-I-IB)}
\begin{itemize}
\item B (``beginning'') marks the first token of a sequence.
\item I (``inside'') continues the sequence within the same line.
\item IB (``inside-below'') continues the sequence on a new line, reflecting multi-line fields.
\end{itemize}

\subsection{Training Strategy}

\subsubsection{Pre-training}
We first convert PubTables-1M data into our required token classification format. Since PubTables-1M contains research documents without label information, the model is pretrained using only two classification heads: Column and Row heads. This pre-training phase focuses on teaching Spatial ModernBERT to identify tabular structures (columns and rows) across diverse table formats found in the PubTables-1M corpus.

\subsubsection{Fine-Tuning}
We subsequently fine-tune the model on our Financial Document Dataset (proprietary), combined with 30\% of the original pre-training data. This approach helps
retain general tabular knowledge while adapting to the specific
distribution of financial documents. Unlike the PubTables-1M dataset, our Financial Document Dataset contains label information (e.g., PO Number, PO Date, Item Description, etc.), which enables us to add the Label head during fine-tuning. Thus, the fine-tuning phase utilizes all three classification heads: Label, Column, and Row.

\subsubsection{Loss Function}
\begin{itemize}
\item We use Cross Entropy Loss for the Label, Column, and Row heads, treating them as multi-class classification tasks.
\end{itemize}

\subsubsection{Optimized Training Components}
Based on our comprehensive ablation study, we identified and implemented a set of critical enhancements that significantly improved model performance:

\begin{itemize}
\item \textbf{Weighted Loss Strategy}: We apply differential weighting to classification heads, allocating 60\% weight to the column head, 30\% to the label head, and 10\% to the row head. This weighting addresses the greater complexity of column structure prediction, leading to more accurate tabular reconstructions.

\item \textbf{Enhanced Spatial Embeddings}: Our final model incorporates both 2D-positional coordinates ($x_{\min}$, $y_{\min}$, $x_{\max}$, $y_{\max}$) and dimensional properties ($width$, $height$) in the spatial embedding, providing richer layout understanding and improving cell boundary detection similar to LayoutLMv3 \cite{layoutlmv3}.

\item \textbf{Auxiliary Bounding Box Regression}: We implement a bounding box regression head that predicts token coordinates as an auxiliary task, providing additional spatial awareness that enhances generalization to unseen document formats.

\item \textbf{Column Consistency Loss}: Our novel column consistency loss ensures structural coherence by minimizing prediction variance across tokens belonging to the same column, significantly improving tabular structure recognition.

\item \textbf{Multi-faceted Data Augmentation}: We employ a combination of spatial noise injection (Gaussian noise to coordinates), text augmentation (dynamic masking, numerical substitution), and layout augmentation (column reordering, random scaling) to enhance model robustness.
\end{itemize}

These optimized components are integrated into our training pipeline, resulting in the state-of-the-art performance demonstrated in our comparative evaluations.

\subsection{Post-Processing}
After obtaining predictions from the heads, we merge tokens using B-I-IB tags to form coherent segments. We then align each segment according to predicted column and row tags. A simple heuristic assigns row numbers to segments that share the same row head label (i.e., B-row, I-row, IB-row) and similarly for headers. For column assignment, we use the column head predictions to determine which column each segment belongs to. Finally, the label head provides the semantic meaning of each segment, identifying the exact type of each column (e.g., Item Description, Quantity, Base Cost) as well as key-value pairs that exist outside tabular structures. This comprehensive approach yields a fully structured representation that can be exported in CSV, JSON, or other tabular formats, with both tabular data and key-value fields properly organized.

\section{Experiments and Results}

\subsection{Datasets}
\begin{itemize}
\item \textbf{PubTables-1M}: A publicly available dataset containing table-annotated documents \cite{pubtables} that we used for our model's pre-training.
\item \textbf{Financial Document Dataset}: Real-world financial documents (invoices, purchase orders, receipts, etc.) annotated with bounding boxes, NER labels (e.g., PO Number, MRP, etc.), column indices, and row types, which we used for model fine-tuning.
\end{itemize}

\subsection{Experimental Setup}

\subsubsection{Implementation}
Spatial ModernBERT is implemented using the Hugging Face Transformers library and PyTorch, with additional layers for spatial embedding and classification extending ModernBERT-base \cite{modernbert}.

\begin{itemize}
\item \textbf{Total Trainable Parameters}: 151,368,795
  \begin{itemize}
  \item ModernBERT-base Parameters: 149,014,272 \cite{modernbert}
  \item Spatial Embedding Parameters: 315,904
  \item Classifier Head Parameters: 2,038,619
  \end{itemize}
\end{itemize}

\subsubsection{Hardware}
Training conducted on standard GPU servers (NVIDIA A10G and L40S) with batch sizes tuned to maximize GPU memory usage.

\subsubsection{Hyperparameters}
\begin{itemize}
\item Learning rate: $1\times10^{-4}$ during pre-training and $5\times10^{-5}$ during fine-tuning, with cosine annealing schedule and warmup steps of 10\% of total training steps.
\item Batch size: typically 32 or 64, depending on GPU capacity.
\item Epochs: 5-10 epochs for pre-training, followed by 10-15 epochs for fine-tuning.
\end{itemize}

\subsubsection{Inference Performance}
We benchmarked the model's performance across various batch sizes and sequence lengths to evaluate scaling properties. The system demonstrates efficient token processing with increasing batch sizes, showing particular optimization at longer sequence lengths. All inference benchmarks were conducted on NVIDIA A10G GPUs to ensure consistent measurements.

\begin{table}
\centering
\caption{Inference Performance Across Different Batch Sizes and Sequence Lengths}
\label{tab:inference}
\begin{tabular}{ccc}
\toprule
Batch Size & Sequence Length & Tokens Per Second \\
\midrule
1 & 128 & 7,301 \\
8 & 128 & 53,637 \\
64 & 128 & 62,329 \\
1 & 256 & 14,511 \\
8 & 256 & 57,048 \\
64 & 256 & 61,951 \\
1 & 512 & 29,402 \\
8 & 512 & 59,409 \\
64 & 512 & 60,985 \\
1 & 1024 & 50,610 \\
8 & 1024 & 59,639 \\
64 & 1024 & 60,212 \\
1 & 2048 & 53,052 \\
8 & 2048 & 58,551 \\
64 & 2048 & 58,527 \\
\bottomrule
\end{tabular}
\end{table}

\begin{figure}[!htbp]
\centering
\includegraphics[width=0.9\columnwidth]{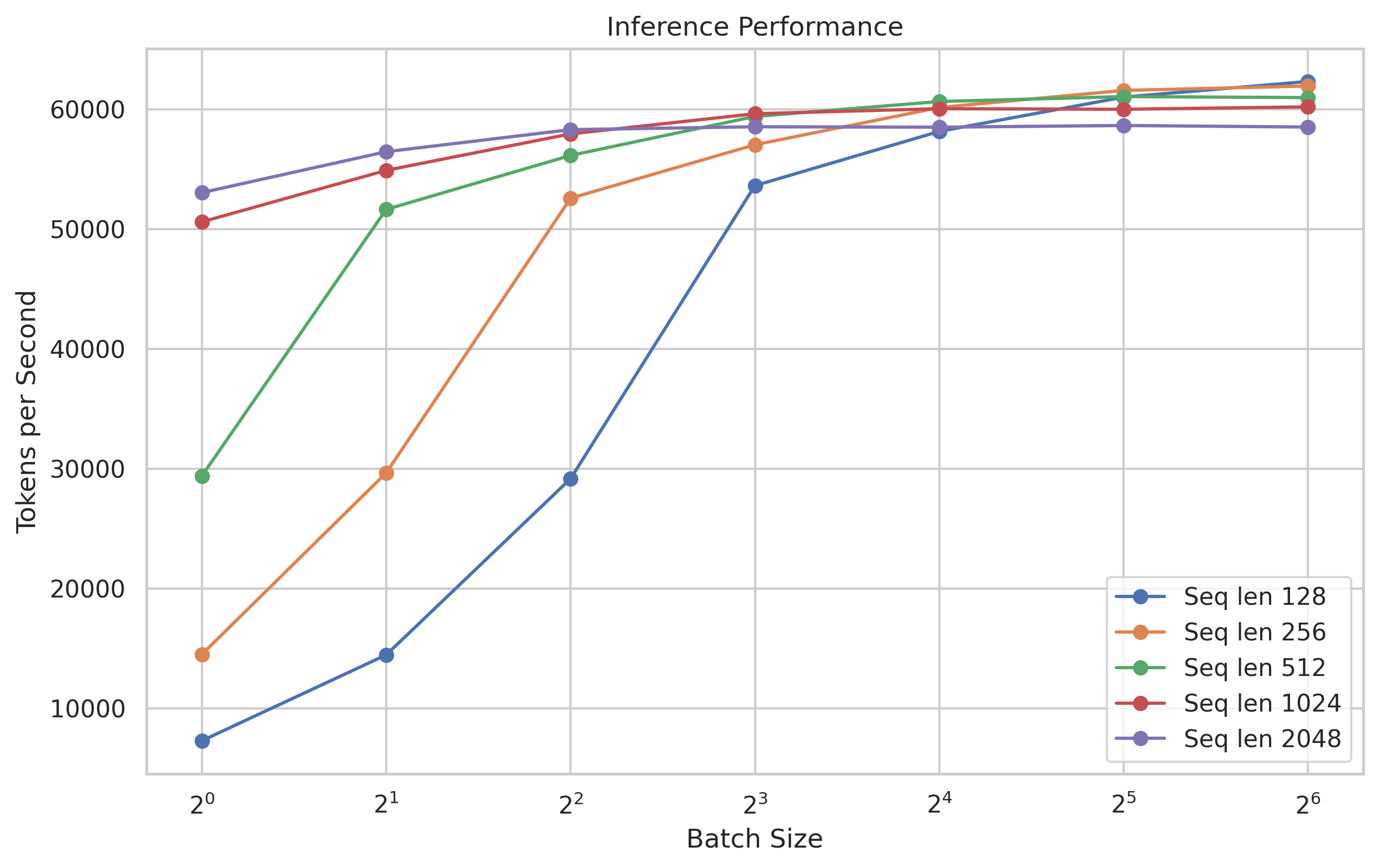}
\caption{Visual representation of the model's inference performance showing tokens processed per second across different batch sizes and sequence lengths.}\label{fig:inference_performance}
\end{figure}

We observe lower token processing rates at shorter sequence lengths (128, 256, 512) for small batch sizes (1-8). This occurs because the model cannot fully leverage the parallelization capabilities of GPUs with these smaller workloads. As sequence lengths and batch sizes increase, we achieve more efficient GPU utilization, reaching maximum throughput of approximately 60,000 tokens per second, as shown in Table~\ref{tab:inference} and Fig.~\ref{fig:inference_performance}.

It's important to note that since Spatial ModernBERT performs token classification rather than generating new tokens, our inference metrics may not be directly comparable to autoregressive models. Token classification using only an encoder model allows for efficient parallel processing of all input tokens simultaneously, unlike autoregressive models which generate tokens sequentially and incur additional computational overhead with each new token.

\subsection{Evaluation Metrics}
We employ a comprehensive evaluation framework to assess Spatial ModernBERT's performance across multiple dimensions of the table and key-value extraction tasks:

\subsubsection{Token-Level Classification Metrics}
For each classification head (Label, Column, and Row), we compute:

\begin{itemize}
\item \textbf{Precision ($P$)}: The proportion of correctly predicted positive instances among all predicted positive instances.
    \begin{equation}
    P = \frac{TP}{TP + FP}
    \end{equation}
    where $TP$ is true positives and $FP$ is false positives.

\item \textbf{Recall ($R$)}: The proportion of correctly predicted positive instances among all actual positive instances.
    \begin{equation}
    R = \frac{TP}{TP + FN}
    \end{equation}
    where $FN$ is false negatives.

\item \textbf{F1 Score}: The harmonic mean of precision and recall, providing a balanced measure of the model's performance.
    \begin{equation}
    F1 = 2 \cdot \frac{P \cdot R}{P + R}
    \end{equation}
\end{itemize}

We calculate these metrics for each class in our multi-class classification tasks and obtain both macro-averaged (treating all classes equally) and weighted-averaged (accounting for class imbalance) scores.

\subsubsection{Table Structure Evaluation}
For evaluating the quality of extracted table structures:

\begin{itemize}
\item \textbf{Fully Correct Table Detection Score}: Percentage of documents where the entire table structure is correctly identified without any structural or content errors.

\item \textbf{Tree-Edit Distance Similarity (TEDS)}: Quantifies structural and textual similarity between predicted and ground-truth tables by computing the minimum-cost sequence of operations (insert, delete, rename) needed to transform one tree into another.
    \begin{equation}
    TEDS = 1 - \frac{EditDistance(T_{pred}, T_{gt})}{max(|T_{pred}|, |T_{gt}|)}
    \end{equation}
    where $T_{pred}$ and $T_{gt}$ are the predicted and ground-truth table structures represented as trees, and $|T|$ denotes the tree size.
\end{itemize}

\subsubsection{Key-Value Extraction Metrics}
For evaluating key-value pair extraction:

\begin{itemize}
\item \textbf{Field Accuracy}: Percentage of correctly extracted key fields.

\item \textbf{Value Accuracy}: Percentage of correctly extracted values for identified keys.

\item \textbf{Exact Match}: Proportion of key-value pairs where both the key and its corresponding value are correctly extracted.

\item \textbf{Levenshtein Similarity}: For value fields, we compute the character-level Levenshtein distance normalized by field length to account for minor OCR or extraction errors:
    \begin{equation}
    Sim_{Lev} = 1 - \frac{Levenshtein(V_{pred}, V_{gt})}{max(|V_{pred}|, |V_{gt}|)}
    \end{equation}
\end{itemize}

For benchmark comparisons with prior work, we primarily use F1 scores for FUNSD and CORD datasets, and TEDS scores for FinTabNet and PubTabNet, as these are the established metrics for these benchmarks.

\subsection{Comparative Benchmark Results}
To evaluate our model's performance against existing state-of-the-art approaches, we conducted experiments on standard benchmarks for document understanding tasks.

\begin{table}
\centering
\begin{threeparttable}
\caption{Comparative performance on FUNSD \cite{funsd} and CORD \cite{cord} benchmarks (F1 scores)}
\label{tab:funsd_cord}
\begin{tabular}{lcc}
\toprule
Model & FUNSD (F1) & CORD (F1) \\
\midrule
BERT-Base \cite{bert} & 60.26 & 89.68 \\
RoBERTa-Base \cite{roberta} & 66.48 & 93.54 \\
BROS-Base \cite{bros} & 83.05 & 95.73 \\
LiLT-Base \cite{lilt} & 88.41 & 96.07 \\
LayoutLM-Base \cite{layoutlm} & 79.27 & - \\
SelfDoc \cite{selfdoc} & 83.36 & - \\
TILT-Base \cite{tilt} & - & 95.11 \\
XYLayoutLM-Base \cite{xylayoutlm} & 83.25 & - \\
LayoutLMv2-Base \cite{layoutlmv2} & 82.76 & 94.95 \\
DocFormer-Base \cite{docformer} & 83.34 & 96.33 \\
LayoutLMv3-Base \cite{layoutlmv3} & - & 96.56 \\
Ours & 73.41 & 95.49 \\
\bottomrule
\end{tabular}
\begin{tablenotes}\small
\item Note: Entries marked with `-' indicate that results were not reported in the original papers or were evaluated on different metrics, making direct comparison not possible.
\end{tablenotes}
\end{threeparttable}
\end{table}

While our model does not achieve the highest F1 score on the FUNSD dataset, it demonstrates competitive performance on the CORD dataset with an F1 score of 95.49, as shown in Table~\ref{tab:funsd_cord}. This suggests that our approach is particularly effective for structured documents like receipts and invoices, which aligns with our target use case of financial document extraction.

The relatively lower performance on FUNSD compared to other models can be attributed to the nature of the dataset and our training strategy. FUNSD primarily consists of form-based documents with key-value pairs rather than tabular structures. Our pre-training focused heavily on table structure recognition (using PubTables-1M), with limited exposure to form understanding during fine-tuning. Since our primary objective was tabular extraction in financial documents, this trade-off was acceptable for our use case, though it highlights the importance of dataset alignment with specific extraction tasks. In future work, we plan to incorporate more key-value extraction data during the pre-training phase to enhance the model's performance on form-based documents while maintaining its strong capabilities for table extraction.

Additionally, we evaluated our model on table structure recognition benchmarks, specifically FinTabNet \cite{fintabnet} and PubTabNet \cite{pubtabnet} validation sets, which focus more directly on our primary task of table structure recognition.

\begin{table}
\centering
\begin{threeparttable}
\caption{Comparative performance on FinTabNet validation set (TEDS \cite{teds} scores)}
\label{tab:fintabnet}
\begin{tabular}{lccc}
\toprule
Model & Simple & Complex & All \\
\midrule
VAST \cite{vast} & - & - & 98.21\% \\
Ly et al. \cite{ly-et-al} & - & - & 95.32\% \\
Ly and Takasu \cite{ly-and-takasu} & - & - & 95.74\% \\
MuTabNet \cite{mutabnet} & - & - & 97.69\% \\
Ours & 98.36\% & 97.78\% & \textbf{98.09\%} \\
\bottomrule
\end{tabular}
\begin{tablenotes}\small
\item Note: Missing values (`-') for Simple and Complex categories occur where prior works only reported aggregate performance across all table types, or where the exact breakdown was not provided in the original publications.
\end{tablenotes}
\end{threeparttable}
\end{table}

\begin{table}
\centering
\begin{threeparttable}
\caption{Comparative performance on PubTabNet validation set (TEDS \cite{teds} scores)}
\label{tab:pubtabnet}
\begin{tabular}{lccc}
\toprule
Model & Simple & Complex & All \\
\midrule
EDD \cite{edd} & - & - & 88.30\% \\
SEM \cite{sem} & - & - & 93.70\% \\
LGPMA \cite{lgpma} & - & - & 94.60\% \\
FLAG-Net \cite{flag-net} & - & - & 95.10\% \\
NCGM \cite{ncgm} & - & - & 95.40\% \\
TableFormer \cite{tableformer} & 95.40\% & 90.10\% & 93.60\% \\
TRUST \cite{trust} & - & - & 96.20\% \\
GridFormer \cite{gridformer} & - & - & 95.84\% \\
VAST \cite{vast} & - & - & 96.31\% \\
Ly et al. \cite{ly-et-al} & 97.89\% & 95.02\% & 96.48\% \\
Ly and Takasu \cite{ly-and-takasu} & 98.07\% & 95.42\% & 96.77\% \\
MuTabNet \cite{mutabnet} & 98.16\% & 95.53\% & 96.87\% \\
Ours & 97.12\% & 96.70\% & \textbf{96.91\%} \\
\bottomrule
\end{tabular}
\begin{tablenotes}\small
\item Note: Empty entries (`-') indicate that the original works did not report separate results for Simple and Complex table categories, instead only provided overall performance metrics.
\end{tablenotes}
\end{threeparttable}
\end{table}

Our model achieves state-of-the-art performance on FinTabNet with a TEDS score of 98.09\%, surpassing most previous methods, as illustrated in Table~\ref{tab:fintabnet}. On PubTabNet, our approach delivers competitive results with a 96.91\% TEDS score, as shown in Table~\ref{tab:pubtabnet}. It should be noted that since our current implementation does not explicitly handle complex tables, we treat complex tables as simple by ignoring their column span and rowspan values and labeling them with their starting column and row indices. For this reason, our results on complex tables may not be directly comparable to other methods that use different approaches to handle merged cells.

\begin{figure*}[t]
\centering
\includegraphics[width=\textwidth]{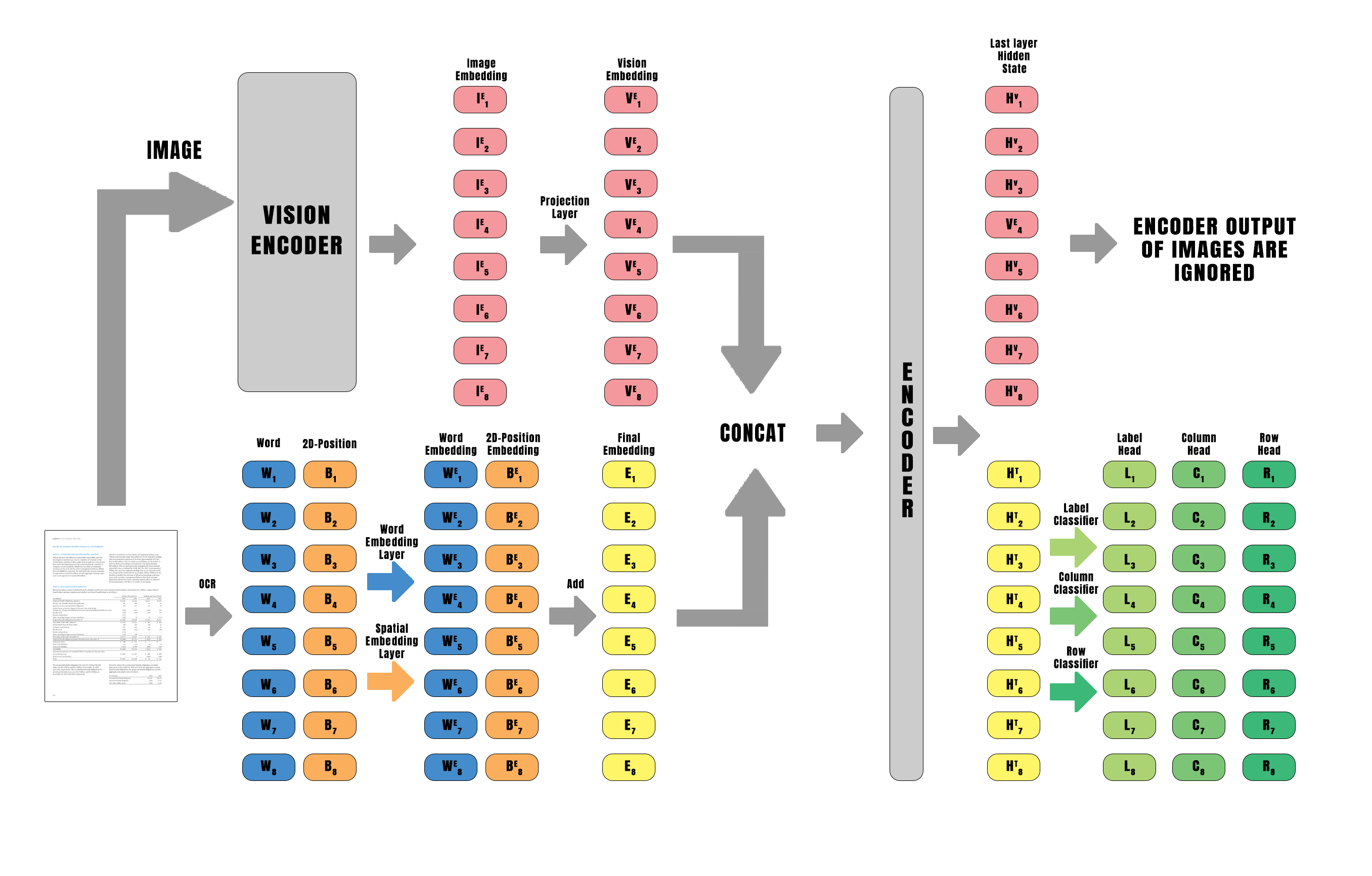}
\caption{Vision-enhanced architecture of Spatial ModernBERT, showing the integration of visual encoder with text and spatial embeddings for multimodal document understanding.}\label{fig:vision_architecture}
\end{figure*}

\subsection{Ablation Study}
To understand the impact of various design choices in Spatial ModernBERT, we conducted a comprehensive ablation study exploring different architectural and training modifications:

\begin{enumerate}
\item \textbf{Loss Weighting}: We investigated assigning different weights to individual classification heads during training, with configurations emphasizing column structure learning (60\% to column head, 30\% to label head, 10\% to row head). Our experiments revealed that the column prediction task was significantly more challenging than label and row prediction tasks based on Token-Level Classification Metrics. This finding informed our weighting strategy, allocating the highest weight to the column head to improve performance on this more difficult aspect of table structure extraction.

\item \textbf{Spatial Embedding Architectures}:
   \begin{itemize}
   \item \textbf{X-Coordinate Emphasis}: Increased dimensionality for horizontal positioning information (340 dimensions each for $x_{\min}$ and $x_{\max}$, 64 dimensions each for $y_{\min}$ and $y_{\max}$). Despite our hypothesis that horizontal positioning might be more informative for tabular structures, this approach did not yield significant performance improvements, leading us to maintain equal dimensionality for all coordinates in our final model.
   \item \textbf{Width and Height Integration}: Incorporated explicit width and height information for each text token (128-dimensional vectors for each). This addition provided modest but consistent performance gains, particularly for identifying cell boundaries and multi-line text spans, and was therefore retained in our final architecture. This is similar to LayoutLMv3 \cite{layoutlmv3}.
   \end{itemize}

\item \textbf{Spatial Position Embedding Initialization}: Implemented sinusoidal initialization inspired by the original Transformer architecture to provide structured spatial position encoding. While sinusoidal initialization showed superior performance during early training epochs, extended training revealed that random normal initialization ultimately achieved better results. This observation suggests that the structured nature of sinusoidal initialization, while beneficial for initial convergence, may constrain the model's ability to learn optimal spatial representations in later epochs.

\item \textbf{Additional Training Objectives}:
   \begin{itemize}
   \item \textbf{Bounding Box Regression}: Introduced an explicit bounding box head, trained with Mean Squared Error loss. This head predicts the normalized coordinates ($x_{\min}$, $y_{\min}$, $x_{\max}$, $y_{\max}$) of each token's bounding box, serving as an auxiliary task to enhance the model's spatial understanding. The regression objective helps the model learn fine-grained spatial relationships between tokens, improving its ability to detect table boundaries and cell alignments. We found that this additional supervision signal particularly enhanced the model's performance on generalized financial documents, i.e., on document formats not found in the training data.
   
   \item \textbf{Column Consistency Loss}: Developed a novel loss that enforces structural consistency in column predictions. Rather than operating on token pairs, this loss function measures the variance in predicted column indices for tokens that belong to the same column according to ground truth. The computation follows these steps:
    \begin{enumerate}
    \item Convert categorical column labels to numerical indices using a mapping function.
    \item Apply softmax to obtain probability distributions over column predictions.
    \item Compute soft column assignments through a weighted sum of probabilities and numerical indices.
    \item For each unique column in the ground truth (excluding padding/outside tokens):
        \begin{itemize}
        \item Identify all tokens belonging to that column
        \item Calculate the variance of their soft column predictions
        \end{itemize}
    \end{enumerate}
    
    The loss is formally defined as:
    \begin{equation}
    L_{consistency} = \frac{1}{B} \sum_{b=1}^B \sum_{c \in C_b} Var(\{\hat{y}_{i,c} | i \in T_{b,c}\})
    \end{equation}
    where $B$ is the batch size, $C_b$ is the set of unique columns in batch $b$, $T_{b,c}$ is the set of tokens belonging to column $c$ in batch $b$, and $\hat{y}_{i,c}$ is the soft column prediction for token $i$. This fully differentiable implementation encourages the model to make consistent column predictions across related tokens while maintaining computational efficiency during training. 
    
    While this approach showed comparable performance on in-domain test sets, it demonstrated significant improvements on out-of-domain documents, where the model previously struggled with column consistency. Specifically, the consistency loss substantially reduced instances where the model would assign different column indices to tokens within the same logical column, greatly enhancing generalization capabilities to previously unseen document formats.
   \end{itemize}

\item \textbf{Data Augmentation Strategies}: 
   \begin{itemize}
   \item \textbf{Spatial Noise Injection}:
      \begin{itemize}
      \item Added controlled Gaussian noise ($\sigma = 5$) to bounding box coordinates (range 0-1000) during training.
      \end{itemize}
   
   \item \textbf{Text Augmentation}:
      \begin{itemize}
      \item Dynamic text masking: Randomly masked 0-20\% of input tokens while preserving their 2D-positional information.
      \item Random token dropout: Independently dropped tokens with 0.1 probability to simulate OCR errors.
      \item Text substitution: Replaced numerical values with similar formats to improve generalization.
      \end{itemize}
   
   \item \textbf{Layout Augmentation}:
      \begin{itemize}
      \item Column reordering: Randomly permuted table columns while maintaining semantic relationships.
      \item Applied random scaling (0.8-1.2) to document dimensions while preserving relative positions.
      \end{itemize}
   
   These augmentation techniques significantly improved model robustness.
   \end{itemize}

\item \textbf{Integration of Visual Information}:
   \begin{itemize}
   \item \textbf{Visual Encoder Architecture}: We incorporated a SigLIP2 \cite{siglip2} visual encoder (google/siglip2-base-patch16-512) to evaluate the potential benefits of visual information, as illustrated in Fig.~\ref{fig:vision_architecture}. The image was processed through the visual encoder to obtain patch embeddings of size $32 \times 32 \times 768$, which were flattened to $1024 \times 768$ dimensions. These embeddings were projected to maintain the same dimensionality ($1024 \times 768$) before being concatenated with text embeddings, with visual embeddings preceding text embeddings in the sequence.
   
   \item \textbf{Basic Training Strategy}: We directly integrated the pretrained SigLIP2 vision encoder into our model architecture as depicted in Fig.~\ref{fig:vision_architecture}. Using the same financial document dataset as used to train our text-only model, we trained the combined architecture with the expectation that the model would implicitly learn to leverage visual cues. During this phase, we kept the vision encoder frozen while allowing other components to be updated through backpropagation.
   
   \item \textbf{Initial Results}: Surprisingly, incorporating visual information without specialized tasks did not improve performance, and in some cases led to a slight decrease in accuracy metrics. We hypothesized that without explicit vision-related tasks or visual document understanding objectives, the model struggled to effectively utilize the visual embeddings alongside textual and spatial information.
   
   \item \textbf{Specialized Training Approaches}: To address this limitation, we experimented with several targeted strategies:
      \begin{itemize}
      \item \textbf{Document-Specific Vision Encoder Pre-training}: We adapted the vision encoder for document understanding by pre-training it on a table-specific object detection task. Specifically, we replaced the backbone in the YOLOX \cite{yolox} architecture with our vision encoder and trained it to identify column boxes before integration into Spatial ModernBERT.
      
      \item \textbf{Word-Patch Alignment Task}: Following an approaches similar to LayoutLMv3 \cite{layoutlmv3}, we implemented a word-patch alignment objective where tokens were randomly masked in the image. A dedicated classification head was trained to predict whether each token is masked or not, creating an explicit incentive for the model to connect visual and textual information.
      
      \item \textbf{Progressive Training Strategy}: We implemented a phased approach beginning with both encoders frozen, training only the projection layer and classifier heads. This allowed visual and textual embeddings to align without destabilizing either representation space, before gradually unfreezing the text encoder for further optimization.
      \end{itemize}
   
   \item \textbf{Final Outcomes}: Despite successful learning of the word-patch alignment task (achieving high classification F1 scores), we observed minimal improvement in our primary table extraction objectives. The addition of visual processing substantially increased computational requirements and training time, without proportional performance gains. Based on this cost-benefit analysis, we decided to exclude vision capabilities from our final model architecture.
   \end{itemize}
\end{enumerate}

Our ablation experiments revealed that combining multiple enhancements — particularly loss weighting, width/height spatial embeddings, bounding box regression, consistency loss, and data augmentations — produced the most robust model for diverse financial document layouts.

\section{Discussion}

Our multi-headed approach, combining label prediction, column classification, and row classification, allows Spatial ModernBERT to effectively capture the 2D structure of financial documents. By leveraging the B-I-IB tagging scheme, the model handles multi-line fields commonly seen in item descriptions or addresses.

\subsection{Advantages of our architecture}

\begin{enumerate}
\item \textbf{Computational Efficiency}: Our architecture is significantly faster and more resource-efficient compared to vision-based and decoder-based approaches, making it ideal for processing large volumes of financial documents.

\item \textbf{Reliability and Accuracy}: Our approach eliminates hallucinations that are common in generative models by treating table extraction as a token classification task rather than a generation task.

\item \textbf{Domain Adaptability}: Our multi-headed approach allows for easy adaptation to new document types through fine-tuning.

\item \textbf{High-Quality Structured Output}: The B-I-IB tagging scheme combined with explicit column and row classification enables precise reconstruction of document structure.

\item \textbf{Robustness to OCR Errors}: The model learns to rely on both textual and spatial context, making it more resilient to OCR errors or poor-quality scans.

\item \textbf{Explainability}: Unlike black-box decoder approaches, our model's predictions are transparent and traceable.

\item \textbf{Low Deployment Overhead}: The model can be deployed as a single inference pipeline without requiring complex post-processing algorithms. Additionally, due to its relatively small model size, it can be efficiently deployed on CPUs while also being able to leverage modern GPU optimizations like flash attention for maximum performance when available.
\end{enumerate}

\subsection{Limitations and Future Work}

Despite strong performance, there are areas for improvement:

\begin{enumerate}
\item \textbf{Multi-Page Documents}: Large invoices spanning multiple pages may require specialized handling and linking of table segments.

\item \textbf{Complex Table Spans}: Some financial documents have nested or merged cells. Additional features (like row-span or col-span) may boost accuracy.

\item \textbf{Zero-Shot Adaptation}: Adapting to previously unseen document layouts or languages remains a challenge. Future work could explore domain adaptation or multilingual training for broader applications.

\item \textbf{Enhanced Key-Value Extraction}: While our current pre-training and fine-tuning phases primarily focus on table structure recognition, future work could explore dedicated pre-training and fine-tuning on larger key-value extraction datasets to improve performance on this complementary task.
\end{enumerate}

\section{Conclusion}

We presented Spatial ModernBERT, a transformer-based system for table and key-value extraction from financial documents. By casting the problem as a token classification task across three heads (label, column, and row) and integrating spatial embeddings directly into the model, we enable robust detection of tabular structures. The empirical results demonstrate state-of-the-art performance on financial document datasets and competitive results on general table extraction benchmarks. Our approach highlights the impact of combining textual and 2D layout signals to address complex document analysis tasks, providing a practical solution for automated information extraction from financial documents. The computational efficiency and accuracy of Spatial ModernBERT make it suitable for real-world deployment in enterprise environments where processing large volumes of diverse financial documents is required.

\bibliographystyle{IEEEtran}

\end{document}